%% file: main.tex
\documentclass[letterpaper,journal]{IEEEtran}
\usepackage{amsmath,amsfonts}
\usepackage{algorithmic}
\usepackage{algorithm}
\usepackage{array}
\usepackage[caption=false,font=normalsize,labelfont=sf,textfont=sf]{subfig}
\usepackage{textcomp}
\usepackage{stfloats}
\usepackage{url}
\usepackage{verbatim}
\usepackage{graphicx}
\usepackage{cite}
\usepackage{array}
\usepackage[table]{xcolor}
\usepackage{xspace}
\usepackage{multirow}
\usepackage{hyperref} 
\usepackage{wrapfig}
\usepackage{booktabs}
\usepackage{pifont}
\usepackage{enumitem}

\begin{document}
\newcommand{\methodname}{PanoLAM\xspace}
\title{\methodname: Large Avatar Model for Gaussian Full-Head Synthesis from One-shot Unposed Image}

\author{Peng Li$^{*}$,
        Yisheng He$^{*\dagger}$,
        Yingdong Hu,
        Yuan Dong,
        Weihao Yuan,
        Yuan Liu, \\
        Siyu Zhu,
        Gang Cheng,
        Zilong Dong,
        Steven Hoi, \textit{Fellow, IEEE}
        and Yike Guo, \textit{Fellow, IEEE}
        \thanks{Peng Li, Yingdong Hu, Yuan Liu, Yike Guo are with the Hong Kong University of Science and Technology, Hong Kong, China (e-mail: plibp@connect.ust.hk; yingdong.hu@connect.ust.hk; yuanly@ust.hk; yikeguo@ust.hk); Yisheng He, Yuan Dong, Weihao Yuan, Gang Cheng, Zilong Dong, and Steven Hoi are with Tongyi Lab, Alibaba Group, China (e-mail: ethanheysh@gmail.com; yuandong15@fudan.edu.cn; whyuan111@gmail.com; chenggang.cg@alibaba-inc.com; list.dzl@alibaba-inc.com; stevenhoi@alibaba-inc.com); Siyu Zhu is with Fudan University, China (e-mail:siyuzhu@fudan.edu.cn); $^{*}$ Equal Contribution. $^{\dagger}$ Corresponding Author
}
}

\maketitle
\begin{figure*}
\hsize=\textwidth
\centering
\includegraphics[width=\textwidth]{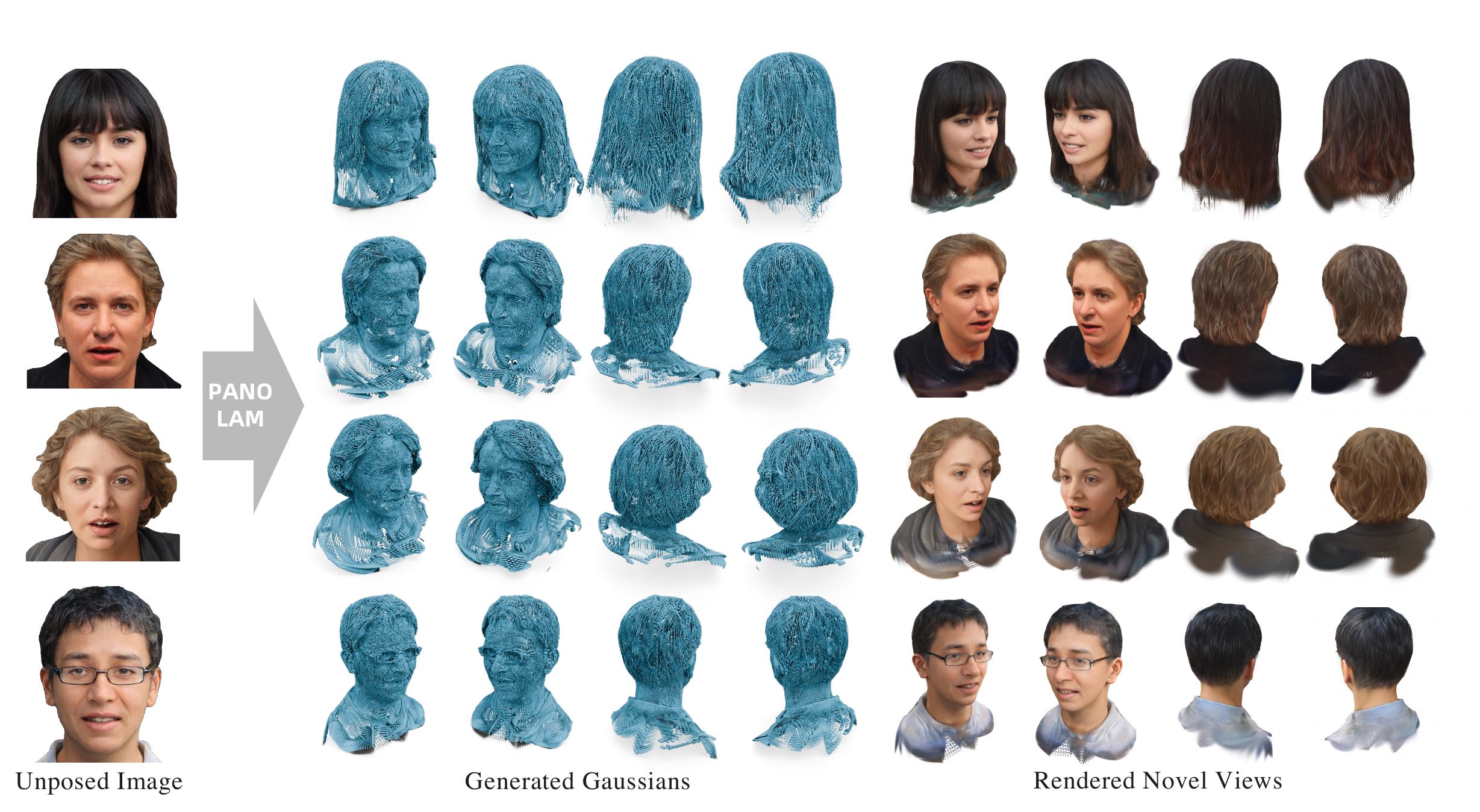}
\caption{PanoLAM creates high-fidelity Gaussian full-heads given an unposed single image in a single forward pass within a second, enabling efficient creation of high-quality 3D avatar assets.}
\label{fig:teaser}
\end{figure*}

\IEEEpubidadjcol

\input{sections/00_abstract}
\input{sections/01_introduction}
\input{sections/02_related_work}
\input{sections/03_method}
\input{sections/04_experiments}
\input{sections/05_conclusion}

\input{main.bbl}

\end{document}

%% file: sections/00_abstract.tex
\begin{abstract}
We present a feed-forward framework for static Gaussian full-head reconstruction and novel-view rendering from a single unposed image. Unlike previous work that relies on time-consuming GAN inversion and test-time optimization, our framework can reconstruct the Gaussian full-head model given a single unposed image in a single forward pass. This enables fast reconstruction and rendering during inference. To mitigate the lack of large-scale 3D head assets, we propose a large-scale synthetic dataset from trained 3D GANs and train our framework using only synthetic data. For efficient high-fidelity generation, we introduce a coarse-to-fine Gaussian head generation pipeline, where sparse points from the FLAME model interact with the image features by transformer blocks for feature extraction and coarse shape reconstruction, which are then densified for high-fidelity reconstruction. To fully leverage the prior knowledge residing in pretrained 3D GANs for effective reconstruction, we propose a dual-branch framework that effectively aggregates the structured spherical triplane feature and unstructured point-based features for more effective Gaussian head reconstruction. Experimental results show the effectiveness of our framework compared to existing work. The dataset is available at \url{https://github.com/PanoLAM/PanoLAM_Dataset}.
\end{abstract}


%% file: sections/01_introduction.tex
\section{Introduction}

\IEEEPARstart{R}{econstructing} head avatars from a single image is an important area of research within computer graphics and computer vision. The capability to generate high-fidelity head avatars opens up new possibilities in areas such as 3D telepresence, video conferencing, filmmaking, gaming, and augmented/virtual reality (AR/VR). However, efficiently creating digital head avatars that are high-fidelity and efficient in rendering is challenging, and various studies have tried to resolve this problem. 

One line of work uses 3D Generative Adversarial Networks (3D-GANs)~\cite{EG3D,panohead} for the unconditional generation of 3D head avatars, which learn 3D generation models from 2D images paired with camera poses. Unlike 2D-GANs~\cite{DBLP:conf/cvpr/StyleGAN} for single-view generation, these approaches can generate face images with control over viewing angles. The pioneering work EG3D~\cite{EG3D} introduces a triplane representation in Cartesian space to enable novel view synthesis after generation. The following works~\cite{panohead,SphereHead} allow a wider range of viewing angles from the generated NeRF space and make 360$^{\circ}$ image synthesis possible. However, these GAN-based frameworks require time- and computation-consuming GAN-inversion~\cite{3DGANINV_KoCCRK23} or test-time optimization~\cite{PTI_RoichMBC23} for image-conditioned generation during inference. Furthermore, the accurate camera pose of the input image is required for joint optimization of these inversion approaches. Rodin~\cite{Rodin_cvpr} and its follow-up~\cite {RodinHD_eccv} instead utilize diffusion models to generate triplanes of the head avatars. However, the multi-step diffusion process during inference still requires minutes of optimization for each case. Moreover, the costly ray marching hinders the rendering resolution of these NeRF-based frameworks. Though 2D super-resolution upsampling is proposed to improve the rendering efficiency and quality, it compromises 3D consistency. 

In this work, we propose a feed-forward framework for static full-head Gaussian reconstruction from a single \textit{unposed} image, enabling fast reconstruction and novel-view rendering at inference time. Achieving this goal presents two main challenges.
\textbf{1)} The first hurdle lies in the lack of large-scale 3D head datasets for network training. To tackle this problem, we propose to leverage the prior knowledge in trained 3D GANs and sample a large-scale, diverse dataset from EG3D~\cite{EG3D} and SphereHead~\cite{SphereHead}. We label and remove invalid cases manually to improve the quality of the dataset.
\textbf{2)} The second challenge involves how to efficiently and effectively reconstruct the high-fidelity Gaussian head from a single unposed image. 

To overcome this, we propose two main designs for our framework.  Firstly, to improve network efficiency, we propose a coarse-to-fine reconstruction mechanism for high-fidelity Gaussian full-head reconstruction. Specifically, sparse points extract rich image features, craft the coarse shape, and are then upsampled for high-fidelity dense Gaussian head reconstruction. By leveraging the topology prior in the FLAME model, we also introduce an efficient network-free upsampling mechanism for features and point clouds densification.
Secondly, to improve the effectiveness of the reconstruction process, we propose a dual-branch framework that hybridizes the unstructured point features and the structured spherical triplane features for the reconstruction of a Gaussian head. To distill the spherical triplane knowledge from the pretrained 3D GAN, we analyze its original feature aggregation manner and propose an efficient mechanism to effectively aggregate multi-layer features from the spherical triplane. The contributions of our work can be summarized as follows:

\begin{itemize}[leftmargin=*]
\item We propose PanoLAM, a large avatar model for \textit{Gaussian full-head} reconstruction from \textit{a single unposed} image.
\item We propose a coarse-to-fine reconstruction mechanism for efficient and high-fidelity Gaussian full-head reconstruction. Built upon the topology prior in the FLAME model, we introduce an efficient network-free upsampling mechanism for points and features densification.
\item We propose a dual-branch framework that hybridizes the representation of points and the spherical triplane prior from 3D GAN. An efficient aggregation mechanism is also proposed to effectively extract multi-layer features from the triplane. 
\item We propose a large-scale, diverse synthesis dataset for 3D avatar reconstruction and generation. To our knowledge, ours is the largest and most diverse 3D head avatar dataset. 
\end{itemize}

%% file: sections/02_related_work.tex
\section{Related Work}
\label{related}

\subsection{3D Head Reconstruction Models}
Various 3D representations have been applied for 3D head modeling, including mesh, Neural Radiance Fields (NeRFs)~\cite{NeRF}, Signed Distance Functions (SDFs), and Gaussian Splatting~\cite{GaussianSplatting}. Mesh-based modelings~\cite{3DMM,BFM,DBLP:journals/tog/LiBBL017} are dedicated to representing 3D human heads with parametric textured meshes and deform the mesh model to improve the topology. NeRF-based works~\cite{Nerfies,DBLP:conf/cvpr/GafniTZN21,AD-NeRF} utilize NeRF for 3D head modeling and get more realistic rendering. To improve geometry quality, SDF-based methods~\cite{ImFacepp_ZhengY0022,PhoMoH_ZanfirAS24} are also introduced, where each 3D position is assigned with the signed distance to the nearest surface. For more efficient rendering, the point-based~\cite{PointAvatar} and Gaussian Splatting-based~\cite{GaussianAvatar,DBLP:conf/cvpr/XuCL00ZL24} frameworks are also proposed. However, these frameworks usually require minutes to hours of optimization from videos or multiview images with estimated camera pose for each person before usage, limiting their capability of scaling up and applications that require fast reconstruction. For general-object reconstruction, M-LRM~\cite{li2025mlrm} uses geometry-aware positional encoding and cross-attention to fuse posed multi-view images. Instead, our work introduces a framework that can generate a Gaussian \textit{full-head} avatar from \textit{a single unposed image} in \textit{a single forward pass} within \textit{seconds}.

\subsection{3D Avatar Generative Models}
One line of generative models utilizes 3D-aware Generative Adversarial Networks (GANs)~\cite{EG3D} to synthesize view-consistent images. Early approaches~\cite{DBLP:conf/iccv/Nguyen-PhuocLTR19,DBLP:conf/3dim/GadelhaMW17} employ explicit 3D representations like meshes and voxel grids, while more recent studies~\cite{EG3D,panohead} utilize implicit representations for better image quality. EG3D~\cite{EG3D} and its follow-ups~\cite{panohead,SphereHead} utilize a 3D triplane representation with GANs to generate heads that are capable of novel view synthesis. However, these methods are mainly designed for unconditional generation and require time- and computation-consuming techniques like GAN inversion~\cite{3DGANINV_KoCCRK23} and test-time optimization~\cite{PTI_RoichMBC23} for image-conditioned generation. The slow optimization procedure during inference and the sacrifice in multi-view consistency hinder their real-world application. ~\cite{DBLP:journals/tog/TrevithickCSCLYKCRN23,TriPlaneNet_BhattaraiNS24} distill an encoding network from EG3D for efficient image-to-triplane generation, but can only generate views near the front face. Instead of using GAN-based frameworks, more recent works~\cite{Rodin_cvpr,RodinHD_eccv} utilize diffusion models to generate triplanes of head avatars. However, the multi-step diffusion process is still slow and computation-intensive during inference. Several works~\cite{GPAAVatar_iclr_ChuLZYLLH24,Next3D_cvpr_SunWWLZZL23,Avat3r_abs-2502-20220,Portrait4D_v2_DengWW24} also introduce generalizable animatable avatar generation, but cannot reconstruct full heads. Moreover, due to the slow rendering speed of NeRF, these approaches require 2D super-resolution networks to enhance image detail, causing view inconsistencies. To enhance rendering quality and speed, recent works~\cite{LAM_Sig25,GAGAvatar,GGHead, li2024era3d,li2025pshuman,chen2025synchuman} also utilize Gaussian Splatting or explicit mesh as the 3D representation for generation. LAM series~\cite{LAM_Sig25,he2026meshlam,hu2026falam} introduces a feedforward framework for single-shot animatable Gaussian/Mesh head generation that can be animated and rendered in real-time on various platforms. However, it requires estimation of FLAME parameters for reconstruction, which is time-consuming, and the generated head can only be rendered from limited viewing angles due to the limited viewing angles in the training 2D videos. Concurrent to this work, recent works like FaceLift~\cite{lyu2025facelift} or PSHead~\cite{yang2025pshead} and~\cite{yang2025single} leverage image or video diffusion priors to enhance the novel view generalization capability. Though effective, the diffusion-based image or video generation process is time-consuming. Instead, this work can reconstruct the 3D \textit{Gaussian full-head} given a single \textit{unposed} image \textit{in a single forward pass} within a second, thanks to our large-scale 3D training datasets and our coarse-to-fine and dual-branch network design. More recently, there are also works like FlexAvatar~\cite{peng2026flexavatar}~\cite{kirschstein2026flexavatar} and UIKA~\cite{wu2026uika} proposed to enhance the viewing angles for animatable complete 3D head reconstruction.

%% file: sections/03_method.tex
\begin{figure*}[t]
    \centering
    \includegraphics[width=1.\linewidth]{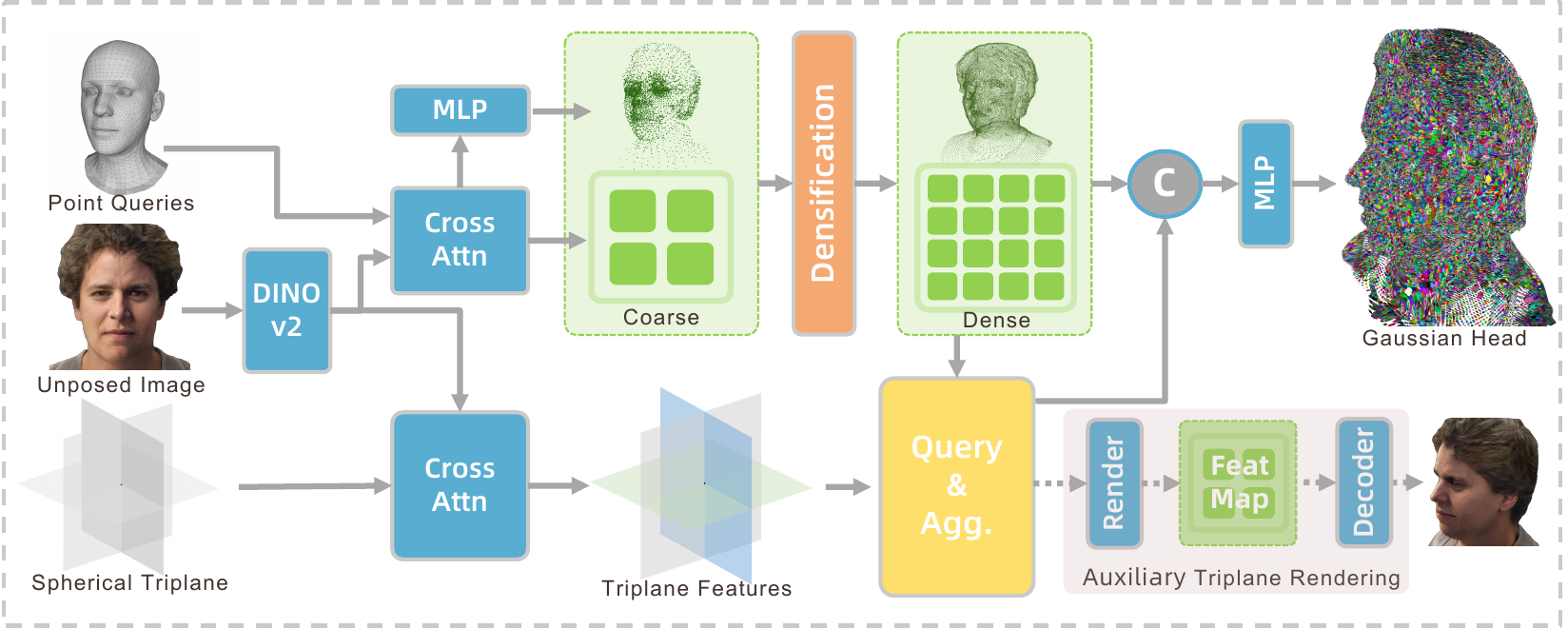}
    \caption{\textbf{Overall Framework.} Given an unposed head image as input, \methodname involves two branches to achieve single-pass 3D Gaussian head reconstruction: a point-based transformer for \textit{coarse-to-fine} point shape reconstruction and point feature extraction, and a spherical triplane transformer to distill prior knowledge from a 3D GAN. Features from the two branches are concatenated for high-fidelity Gaussian head regression.}
    \label{fig:framework}
    \vspace{-4mm}
\end{figure*}

\section{Methodology}

\subsection{Overview}
As shown in Fig.~\ref{fig:framework}, given an unposed reference image, our framework generates the Gaussian full-head in an efficient coarse-to-fine and joint point-triplane manner, where dense point clouds and point features are densified from sparse ones and fused with the spherical triplane features for dense Gaussian head regression. As shown in the upper left part of Fig.~\ref{fig:framework}, in the coarse stage of the point branch, sparse vertices from the FLAME vertices interact with the extracted multi-level image features by stacked cross-attention modules for feature extraction and coarse Gaussian point reconstruction. The dense reconstruction stage leverages barycentric weights from FLAME subdivision to densify the coarse Gaussian point position and features for dense Gaussian point regression. To fully leverage the prior knowledge residing in the spherical triplane of 3D GAN, as shown in the bottom branch of Fig.~\ref{fig:framework}, a spherical triplane branch is also introduced, and a novel efficient multi-layer aggregation mechanism is proposed to aggregate the structured features from the spherical triplane using the densified point cloud, which are then concatenated with the unstructured point features for high-fidelity Gaussian head reconstruction. To optimize the spherical-triplane branch and distill the prior knowledge, in the bottom right of the framework, a frozen SphereHead decoder is utilized to decode the features to images for supervision during training.

\subsection{Coarse-to-fine Gaussian Head Avatar Generation}
\label{sec:CanonicalGSGen}
The number of Gaussian points is one key factor that can affect the fidelity of the reconstructed Gaussian head, especially for the head avatar that contains sharp details like hair strands, mustaches, and wrinkles. The total number of Gaussian points should be large enough to produce these details. However, the large number of Gaussian points is hard to optimize during training, especially in our feed-forward pipeline, where the iterative densification operation in the traditional 3D Gaussian Splatting training pipeline is not applicable. To resolve this problem, unlike previous works~\cite{LAM_Sig25,TripMeetGS_ZouYGLLCZ24} that directly regress the dense point cloud, we introduce a coarse-to-fine mechanism. Specifically, sparse Gaussian points are first reconstructed for coarse shape reconstruction and then densified to dense Gaussians for high-fidelity details crafting. There are two advantages in the proposed design. Firstly, the sparse number of points enables efficient cross-attention with the image features for texture and shape features extraction, saving large computation and memory costs compared with a large number of points. Secondly, the sparse points can be well supervised and optimized for large deformation to carve the shape of the head, especially in regions that require large deformation from the original FLAME model, e.g., long hair. Dense Gaussian points upsampled from these correctly deformed sparse points can focus more on texture details crafting, which are much easier to optimize.

\paragraph{\textbf{Coarse Gaussian Head Generation}}
To fully leverage the head shape prior that resides in the FLAME model, we initialize the coarse Gaussian point with sparse vertices (5,023 points) on the neutral FLAME model in the canonical space. Our target is to regress the deformation offset for shape refinement of regions that FLAME cannot model, e.g., long hairs, glasses, and caps. To supervise the shape deformation with RGB input images, Gaussian attributes for each point are also regressed for differentiable Gaussian Splatting. To get started, we assign positional embedding to each point and utilize learnable multi-layer perceptron (MLP) modules to project the channel of features into the token channel $C_t$ of transformers as:
    $F_{P_0} = MLP(\gamma(V)),$
where $V$ is the spatial position of each vertice and $\gamma$ the L$\_$frequency sinusoidal encoding as in NeRF. To extract the texture and shape information in the given unposed image, we utilize the pre-trained DinoV2~\cite{DinoV2} for feature extraction. Then, we fuse features derived from both shallow and deep layers to obtain both local and global image features $F_{I}$. Specifically, an MLP fuses the $\{5,12,18,24\}$ layers of features in DinoV2 into $C_t$ token channels of features as:
\begin{equation}
    F_I = MLP(\mathcal{C}(F_{D_5}, F_{D_{12}}, F_{D_{18}}, F_{D_{24}}),
    \label{eqn:dpt_dino}
\end{equation}
where $F_{D_{i}}$ is the $i_{th}$ layer of DinoV2 features and $\mathcal{C}$ is the concatenation operation.

Given mapped point features $F_P$ and extracted image features $F_I$, we utilize $L_{\mathcal{A}}$ layers of stacked cross-attention modules $\mathcal{A}=\{\mathcal{A}_{i}\}_{i=1}^{L_{\mathcal{A}}}$ from Transformer for feature extraction from image to point cloud as follows:
\begin{equation}
    F_{P_{i}} = \mathcal{A}_{i}(F_{P_{i-1}}, F_I),
    \label{eqn:attn}
\end{equation}
where $F_{P_{i}}$ is the $i_{th}$ layers of point features in the stacked cross-attention layers.

Following the feature extraction process, each point retains its distinct features. Using these features, we develop decoding headers $\mathcal{D}$, composed of multilayer perceptrons (MLPs), to predict the deformation offset $O_k \in \mathbb{R}^3$ to refine the individual's detailed shape. Gaussian attributes for each point are also regressed for rendering, including color $c_k \in \mathbb{R}^3$, opacity $o_k \in \mathbb{R}$, scale factors $s_{k} \in \mathbb{R}^3$, and rotation $R_k \in SO(3)$. This decoding process is as follows:
\begin{equation}
    \{c_k, o_k, s_k, R_k, O_k\}^{M_C}_{k=1} = \mathcal{D}(F_{P_{k}}),
    \label{eqn:gs_decoder}
\end{equation}
where $M_C=5023$ represents the total number of Gaussians in the coarse reconstruction stage, and $F_{P_{k}}$ denotes the extracted point feature for point $P_k$. Although predicting deformation offsets for each point can enhance the shape, freely moving these points may also negatively impact reconstruction results. We therefore restrict the range of deformation offset to be within $[-\epsilon_{O_C}, \epsilon_{O_C}]$.

\paragraph{\textbf{Efficient Point Cloud Densification}}
After the reconstruction of the coarse Gaussian head, we obtain a sparse point cloud that describes the shape of the target head avatar in the image; each point is also attached with rich features extracted from the image. We then densify these points and features for high-fidelity dense Gaussian head reconstruction. Unlike previous methods~\cite{PUDF_LiLHF21,PU_Net} that require a network trained to upsample sparse points, we propose to leverage the topology in the FLAME mesh for efficient densification. Specifically, the proposed strategy utilizes the barycentric coordinates from mesh subdivision to densify the point cloud and features. Our key observation is that after the deformation in the coarse reconstruction stage, the topology between each point mostly remains as the original FLAME vertices. Thereafter, we can precompute the barycentric coordinates for point cloud densification utilizing mesh subdivision on FLAME. Specifically, given the original FLAME model, we perform subdivision for faces that are larger than a threshold $\lambda_{area}$, the position of newly added vertices $v'$ can be computed from the three vertices of the original face with barycentric coordinates as: $v' = \sum_{i=1}^{3}({w_i \cdot v_i}),$
where $w_i$ and $v_i$ are the barycentric weight and vertex coordinate of the original face, respectively. After the deformation in the coarse reconstruction stage, each vertex moves to a new position w.r.t the predicted offset as $p_i = v_i + O_i$. Then new added point $p'$ with its attached point features $p_{F}'$ can be interpolated with barycentric weights as:
    $p' = \sum_{i=1}^{3}({w_i \cdot p_i})$, and
    $p_{F}' = \sum_{i=1}^{3}({w_i \cdot p_{F_{i}}}),$
where $p_{F_{i}}$ is the extracted features from the last cross-attention blocks in Equ.~\ref{eqn:attn}. In this way, we can efficiently densify the points and features.

\paragraph{\textbf{Dense Gaussian Head Generation.}} After getting the dense point position with dense point features, we can utilize them for dense Gaussian head regression. Rather than using only unstructured point features for prediction, we also aggregate multi-layer features from the structured spherical triplane for more effective Gaussian attributes regression, which we will discuss in Sec.~\ref{sec:spherecal_triplane_distillation}. Denote $F_{P_{i}}$ as the $i_{th}$ point feature and $F_{T_i}$ as the aggregated spherical triplane feature for point $P_i$, the dense Gaussian head attributes and a small deformation residual can be regressed similarly to the coarse one as:
\begin{equation}
    \{c_i, o_i, s_i, R_i, O_i\}^{M_D}_{i=1} = \mathcal{D}(\mathcal{C}(F_{P_{i}}, F_{T_{i}})),
    \label{eqn:dense_gs_decoder}
\end{equation}
where $\mathcal{C}$ denotes the concatenation operation, $M_D$ the number of dense Gaussian points, and $O_i$ the deformation residual restricted within $[-\epsilon_{O_D}, \epsilon_{O_D}]$.

\begin{figure}
\centering
    \includegraphics[width=.9\linewidth]{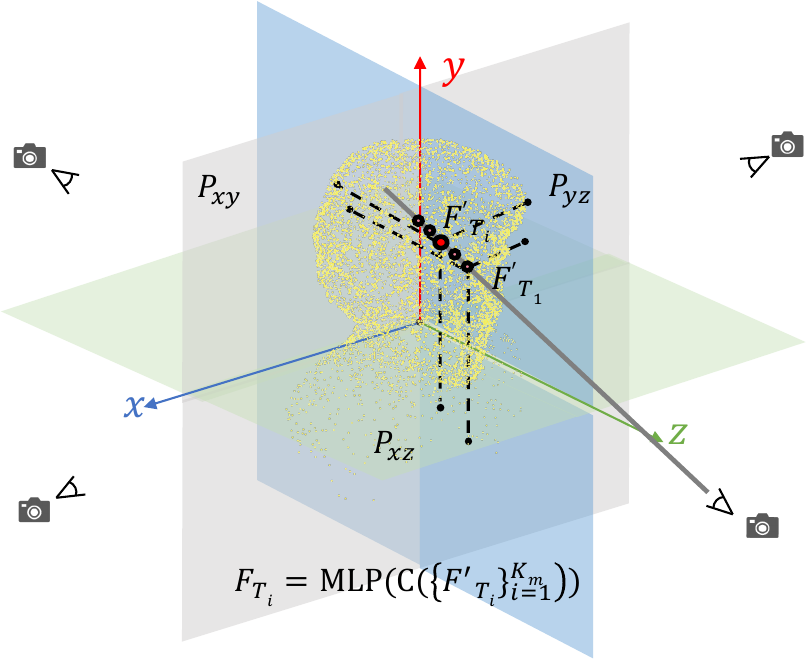}
    \vspace{-2mm}
    \caption{The proposed feature aggregation mechanism samples and aggregates multi-layer features from the spherical triplane for each point.}
    \label{fig:framework_feat_agg}
\end{figure}

\subsection{Spherical Triplane Prior Knowledge Distillation}
\label{sec:spherecal_triplane_distillation}
In the previous section, we obtain the dense point cloud with its attached point features extracted from the unposed image. To better leverage the prior knowledge within SphereHead's framework, we introduce a spherical triplane branch to distill the knowledge from a pretrained model. 
\paragraph{\textbf{Preliminary: The Spherical Triplane representation}} from SphereHead~\cite{SphereHead} combines the three triplanes in the Cartesian coordinate system with another three spherical planes in the spherical coordinate system, where features are queried with projection operations in the respective coordinate systems. This representation alleviates the feature entanglement issue in regular triplanes.

In the spherical triplane branch, we initialize the $H_T \times W_T \times 6 \times C_t$ spherical triplane features with learnable query features, which are then flattened into token features $F_T \in \mathbb{R}^{N_T \times C_t}$, where $N_T = H_T \times W_T \times 6$. These learnable features then interact with the image features $F_I$ with stacked cross-attention blocks  $\mathcal{A_T}=\{\mathcal{A_T}_{i}\}_{i=1}^{L_{\mathcal{A_T}}}$ similar to the learnable point features as:
\begin{equation}
    F_{T_{i}} = \mathcal{A_T}_{i}(F_{T_{i-1}}, F_I),
    \label{eqn:attn_tri}
\end{equation}
where $F_{T_{i}}$ is the $i_{th}$ layer of spherical triplane features in the stacked cross-attention layers. The final output features are then projected by MLPs to the same channel as the spherical triplane in SphereHead. 
We then need to query the feature from the spherical triplane to enhance the point features. 
One vanilla way is to utilize the point-based query strategy that fetches each point's spherical-triplane feature by projection. However, we find that our point clouds are distributed on the shape surface similar to a mesh, and such a single-layer query strategy cannot fully aggregate the corresponding features of each point from the spherical triplane. 

That's because the triplane designed for NeRF-based rendering requires ray marching to aggregate multiple points(/layers) of features for rendering. Using a single point cloud on the shape surface cannot fully fetch valuable features from the spherical triplane. 
While ray marching is computationally consuming and inefficient, we propose to sample multiple layers of points from the single-layer point cloud for feature aggregation. As shown in Fig.~\ref{fig:framework_feat_agg}, we sample 4 virtual cameras that can capture the front, left, right, and back sides of the head. For each camera, we cast a ray from the camera center to each point $P_i$, then $K_m$ points $P'=\{P'_i\}_{i=1}^{K_m}$ are sampled along the ray near the point. Each point then queries features from the spherical triplane by projection. We then concatenate and aggregate these queried features with MLPs to obtain the final spherical triplane features as:
\begin{equation}
    F_{T_i} = MLP(\mathcal{C}(\{F'_{T_i}\}_{i=1}^{K_m})),
    \label{eqn:triplane_feat_agg}
\end{equation}
where $F'_{T_i}$ is the spherical triplane feature queried by sampled point $P'_i$ ; $\mathcal{C}$ is the concatenation operation; and $F_{T_i}$ is the final aggregated feature for point $P_i$. In this way, each point gets its aggregated spherical triplane feature. These aggregated spherical triplane features can be concatenated with the point feature for dense Gaussian attribute regression as in Equ.~\ref{eqn:dense_gs_decoder}. To supervise this branch in distilling knowledge from SphereHead, we add a frozen decoder from SphereHead to decode the feature into the target image, supervised by ground truth images during training, as shown in the bottom right part of Fig.~\ref{fig:framework}.

\subsection{Large-scale Training Data Generation}

\begin{table}[t]
    \caption{Comparisons of existing 3D head datasets. $^\ast$ denotes 3D head with rough captured 3D hair shape.}
    \vspace{-3mm}
    \centering
    \setlength{\tabcolsep}{1.5pt}
    \scriptsize
    \begin{tabular}{@{}lcccllccc@{}}
    \cmidrule[\heavyrulewidth](r){1-4} \cmidrule[\heavyrulewidth](l){6-9}
    \multicolumn{1}{l}{Dataset} & Sub. & Range & Hair         &  & \multicolumn{1}{l}{Dataset} & \multicolumn{1}{c}{Sub.} & \multicolumn{1}{c}{Range} & \multicolumn{1}{c}{Hair}                 \\ \cmidrule(r){1-4} \cmidrule(l){6-9} 
    BU-3DFE                     & 100  & front & \ding{55}    &  & HeadSpace                   & 1519                     & $270^{\circ}$             & \ding{55}         \\ 
    BU-4DFE                     & 101  & front & \ding{51}    &  & FaceScape                   & 938                      & $360^{\circ}$             & \ding{55}         \\
    BJUT-3D                     & 500  & front & \ding{55}    &  & AVA-256                   & 256                     & $360^{\circ}$             & \ding{51}         \\
    Bosphorus                   & 105  & front & \ding{55}    &  & FaceVerse$^\ast$         & 128                      & $360^{\circ}$             & \ding{51}         \\
    FaceWarhouse                & 150  & front & \ding{51}    &  & RenderMe360$^\ast$       & 500                      & $360^{\circ}$             & \ding{51}         \\
    4DFAB                       & 180  & front & \ding{55}    &  & SynHead100                  & 100                      & $360^{\circ}$             & \ding{51}         \\
    BP4D-S                      & 41   & front & \ding{51}    &  & \textbf{PanoLAM-Front}         &   48,135                       & $72^{\circ}$              & \ding{51}         \\
    Nersemble                   & 222  & front & \ding{51}    &  & \textbf{PanoLAM-360$^\circ$}           &   117,186      & $360^{\circ}$             & \ding{51}         \\ 
    \cmidrule[\heavyrulewidth](r){1-4} \cmidrule[\heavyrulewidth](lr){6-9}
    \end{tabular}
    \label{tab:data}
\end{table}

The scale and diversity of training data are critical to the generalization of feed-forward 3D avatar models. As summarized in Table~\ref{tab:data}, existing 3D head datasets are still limited in subject diversity and view coverage, which constrains model robustness under unconstrained real-world inputs. Following prior evidence that models trained on high-quality synthetic data can transfer well to real scenarios~\cite{fakeIt_wood2021fake}, we construct a large-scale training corpus by sampling from two complementary pretrained 3D GAN priors: EG3D for multiview near-frontal synthesis and SphereHead for full $360^{\circ}$ coverage. This hybrid sampling strategy combines their distinct data distributions (e.g., hairstyle and accessory diversity from EG3D, complete view coverage from SphereHead), yielding broader appearance and viewpoint support than using either prior alone.

Concretely, for EG3D we sample within an approximately $72^{\circ}$ view range (pitch $\pm 26^{\circ}$ and yaw $\pm 36^{\circ}$ relative to the frontal direction), and randomize camera radius and focal length to improve adaptation to heterogeneous capture conditions; the radius is sampled from $\mathcal{N}(2.7, 0.1^2)$ and the focal length from $\mathcal{N}(18.83, 1^2)$. Each triplane contributes 32 randomly sampled views. For SphereHead~\cite{SphereHead}, we generate full-head data by combining 8 evenly spaced equatorial views (for guaranteed coverage) with 24 additional random views under the same camera randomization strategy. Since raw GAN samples may contain geometric or appearance artifacts that are harmful for supervised learning, we further build an annotation tool to manually filter out failure cases. After quality control, the final datasets contain 48,135 subjects for PanoLAM-Front and 117,186 subjects for PanoLAM-$360^\circ$. 

\subsection{Optimization and Regularization}
In the training phase, we randomly select $N_v$ different views of images of the same subject. We randomly choose one as the reference image for the Gaussian full-head reconstruction, while the others serve as novel view images for prediction. Note that both branches in our framework render images for supervision. We ensure the accuracy of the rendered RGB images from each branch by comparing them with the ground truth target images, utilizing a combination of $\mathcal{L}_1$ loss and perceptual loss for supervision:
\begin{equation}
    \mathcal{L}_{rgb} = \lambda_{1}\mathcal{L}_1 + \lambda_{2}\mathcal{L}_{lpips}.
    \label{eqn:loss_rgb}
\end{equation}
Additionally, for the sparse and dense Gaussian Splatting branch, we render the silhouette and supervise it using $\mathcal{L}_1$ loss, referred to as $\mathcal{L}_{mask}$. For the spherical triplane branch, we supervise the extracted feature with those in the pretrained SphereHead by $\mathcal{L}_1$ loss, denoted as $\mathcal{L}_{st}$. The total loss function is a weighted sum as:
\begin{equation}
    \mathcal{L} = \lambda_3\mathcal{L}_{rgb}^{G} + \lambda_4\mathcal{L}_{rgb}^{ST} + \lambda_5\mathcal{L}_{mask} + \lambda_6\mathcal{L}_{st},
\end{equation}
where $\mathcal{L}_{rgb}^{G}$ denotes the RGB image loss for sparse and dense Gaussian Splatting and $\mathcal{L}_{rgb}^{ST}$ for the auxiliary spherical triplane branch, both denoted in Equ.~\ref{eqn:loss_rgb}.

%% file: sections/04_experiments.tex
\section{Experiments}

\begin{figure*}
  \centering
    \footnotesize\leftline{~~~~\qquad\qquad\qquad\qquad\qquad~~~Synthesis dataset~~~~~\qquad\qquad\qquad\qquad\qquad\qquad\qquad\qquad\qquad\qquad\qquad~~~~In-the-wild dataset~~~}\includegraphics[width=1.\linewidth]{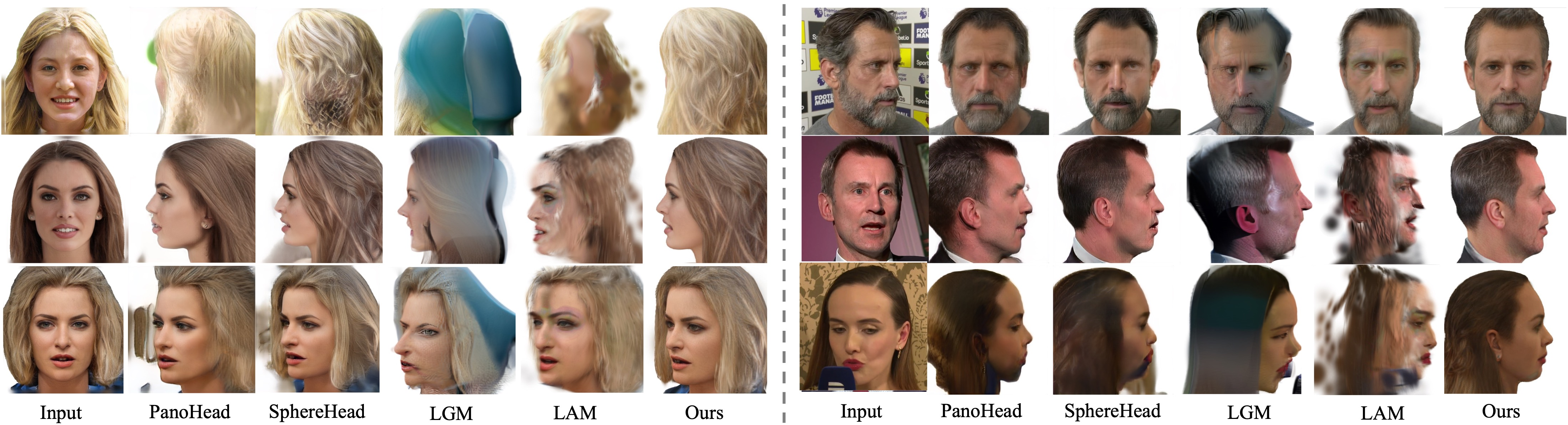}
    
  \caption{Visualization of reconstruction and novel view synthesis of different methods.  The left part shows comparisons of different methods on synthesis data, and the right part shows visualization on in-the-wild real-world images, showing the out-of-domain generalizability of PanoLAM trained on our curated datasets.}
  \vspace{-4mm}
  \label{fig:vis_test}
\end{figure*}

\begin{figure*}
  \centering
    \includegraphics[width=\linewidth]{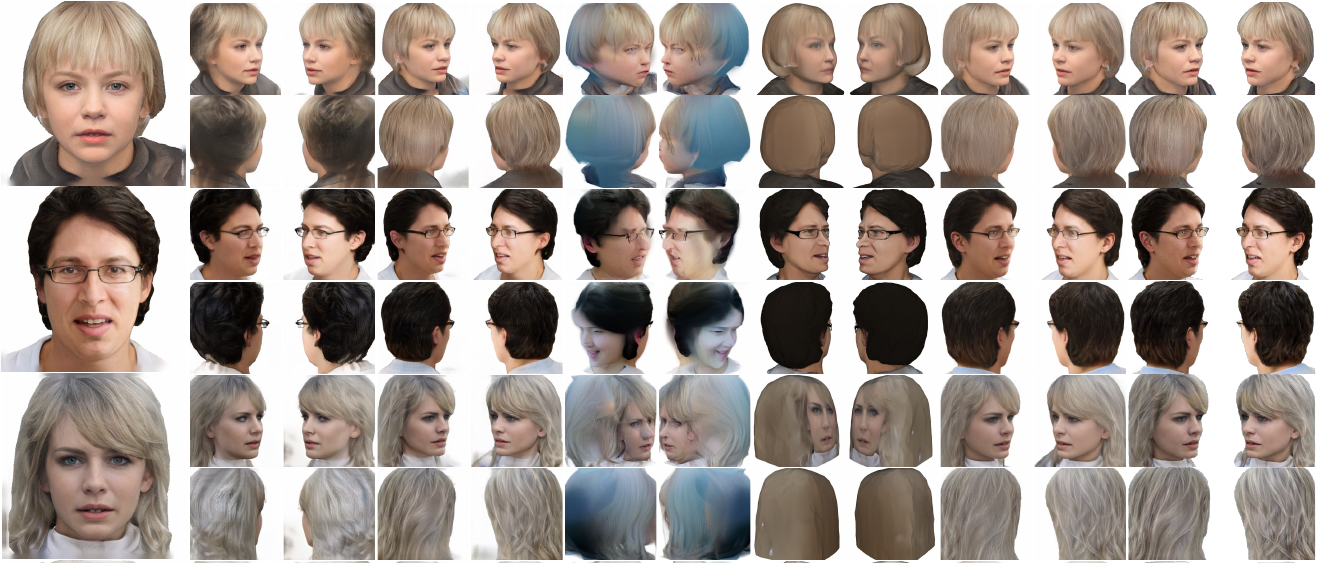}
    \footnotesize\leftline{~~~\qquad~~~Input~~~\quad\qquad\qquad~PH-PTI~~~\qquad\qquad~SH-PTI~\qquad~\qquad\qquad~LGM~\qquad\qquad~ Hunyuan3D-2~\qquad\qquad\quad~Ours~~~\qquad\qquad\qquad~GT}
  \caption{Comparison of reconstruction and novel view synthesis of different methods on the test set. Our framework can reconstruct 360-degree heads given a single input view, with fine-grained details and fewer artifacts compared to baselines.}
  \label{fig:more_cmp_testset}
\end{figure*}

\begin{figure}[t]
    \centering
    \includegraphics[width=\linewidth]{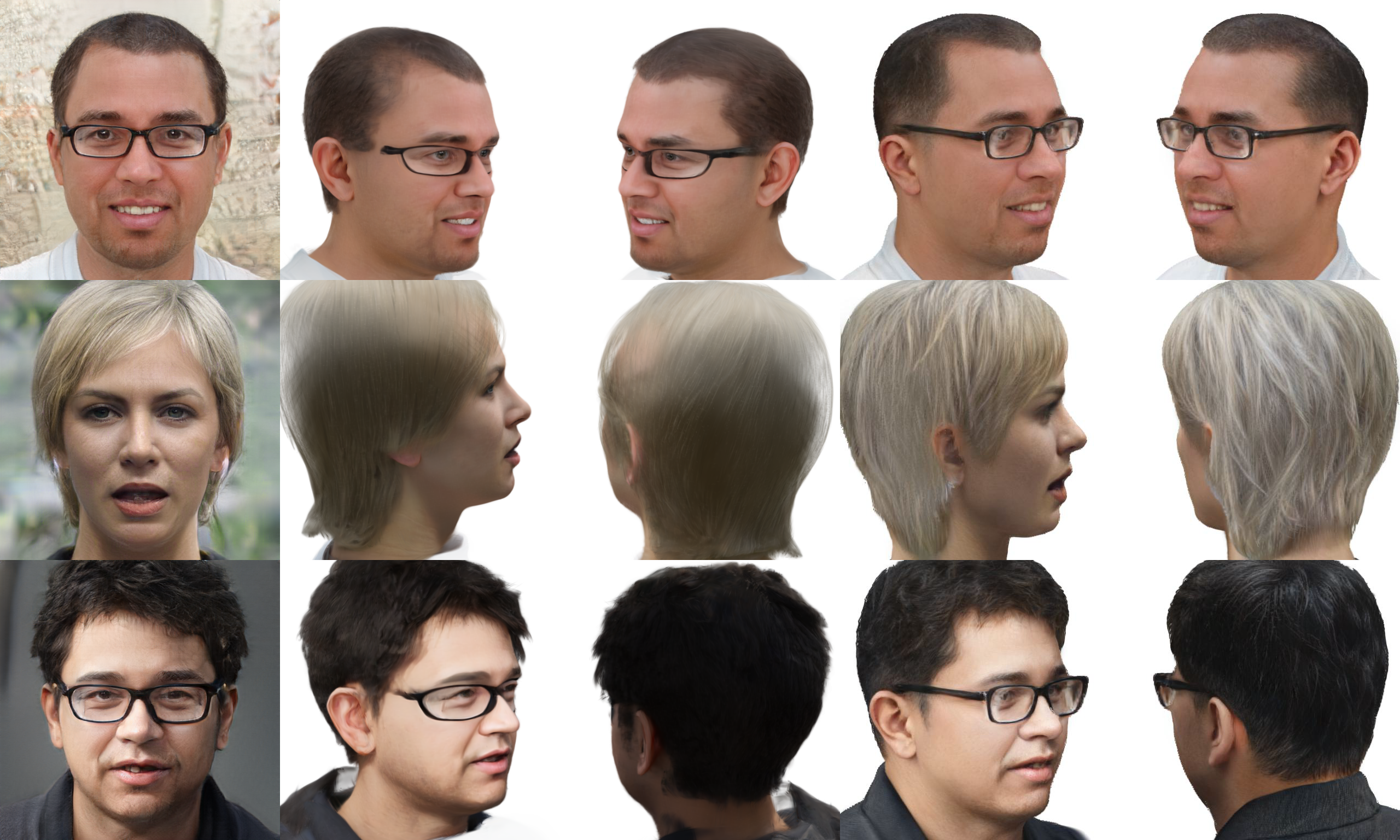}
    \vspace{-5pt}
    \small\leftline{~\quad~Input~~~~~~\qquad~FaceLift~~~~~~\quad\qquad~~\qquad~~Ours}
    \caption{Qualitative comparisons with FaceLift~\cite{lyu2025facelift}.
    }
    
    \label{fig:comp_facelift}
\end{figure}

\subsection{Experiments Setting}
\paragraph{Implementation Details}
Our framework is implemented in PyTorch, with the DINOv2 image feature extraction backbone frozen. The Transformer comprises 8 layers of basic blocks, featuring 16 attention heads and $C_t=512$ feature dimensions. The extracted features are translated into Gaussian attributes through three multilayer perceptron (MLP) layers. The network is trained over 100 epochs using the Adam optimizer and a cosine warm-up scheduler. Hyperparameters are empirically set as $N_{v}=6$, $\lambda_{1}=\lambda_{2}=1.0$, $\lambda_{3}=1.0$, $\lambda_{4}=0.1$, $\lambda_{5}=1.0$, $\lambda_{6}=0.0001$, $\epsilon_{O_C}=0.15$, $\epsilon_{O_D}=0.056$, $M_D=10K$, $K_m=32$, $\lambda_{area}=2.0e-6$, $H_T=W_T=64$.

\paragraph{Datasets}
We employ the large-scale dataset that we have generated for training. To assess the 360$^\circ$ synthesis capability, we randomly select 100 subjects from our PanoLAM-360$^\circ$ dataset for testing, and the remaining for training. 

\paragraph{Baseline fine-tuning}
We fine-tune LGM and LAM on PanoLAM-Front and PanoLAM-$360^\circ$ using the same split, excluding all views of the 100 test identities. LGM is initialized from its official checkpoint and trained with four $256\times256$ input views and eight $512\times512$ supervision views per identity, with ImageDream frozen. LAM is initialized from LAM-20K and trained with one $504\times504$ source image and eight $512\times512$ supervision views; source-view FLAME parameters are fitted with VHAP, while DINOv2, FLAME, and the fitted parameters remain frozen. Both use AdamW for 20,000 steps with an effective batch size of 32, 1,000 warm-up steps, and cosine decay to $10^{-6}$; the initial learning rates are $5\times10^{-5}$ for LGM and $10^{-5}$ for LAM.

\paragraph{Evaluation Metrics}
We evaluate the quality of our synthesized images using three quantitative metrics: Peak Signal-to-Noise Ratio (PSNR), Structural Similarity Index (SSIM), and Learned Perceptual Image Patch Similarity (LPIPS).

\paragraph{FaceLift inference}
For FaceLift, we use the official pretrained models and inference pipeline, with a single source image as input and no additional ground-truth views. We retain the default preprocessing and inference settings: automatic face cropping, an input resolution of $512\times512$, 50 diffusion steps, a guidance scale of 3.0, and a random seed of 4. The multi-view diffusion model generates six views, which are subsequently passed to GS-LRM to reconstruct the Gaussian head.

\subsection{Main Results}
\begin{figure}[t]
    \centering
    \small\subfloat[Thick glasses.]{%
        \includegraphics[width=0.49\linewidth]{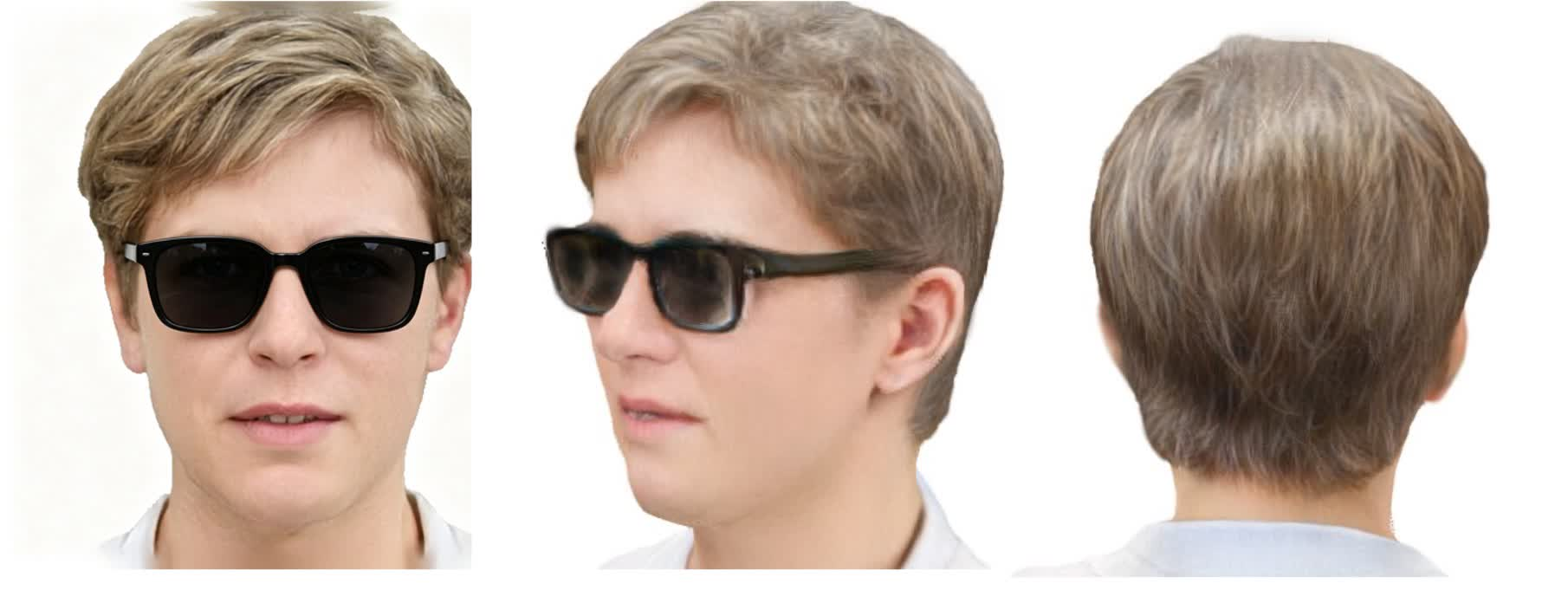}%
    }\hfill
    \small\subfloat[Partial hand occlusion.]{%
        \includegraphics[width=0.49\linewidth]{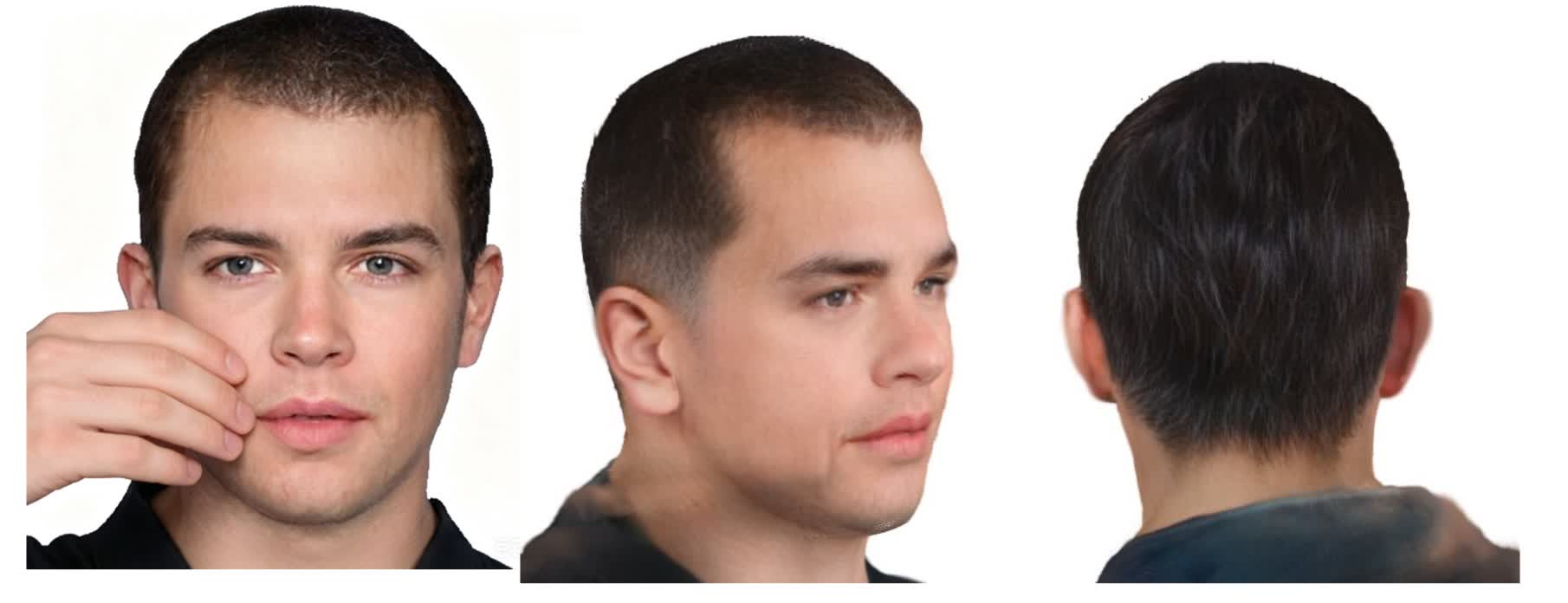}%
    }
    \vspace{0.1pt}
    \small\subfloat[Extreme light condition.]{%
        \includegraphics[width=0.49\linewidth]{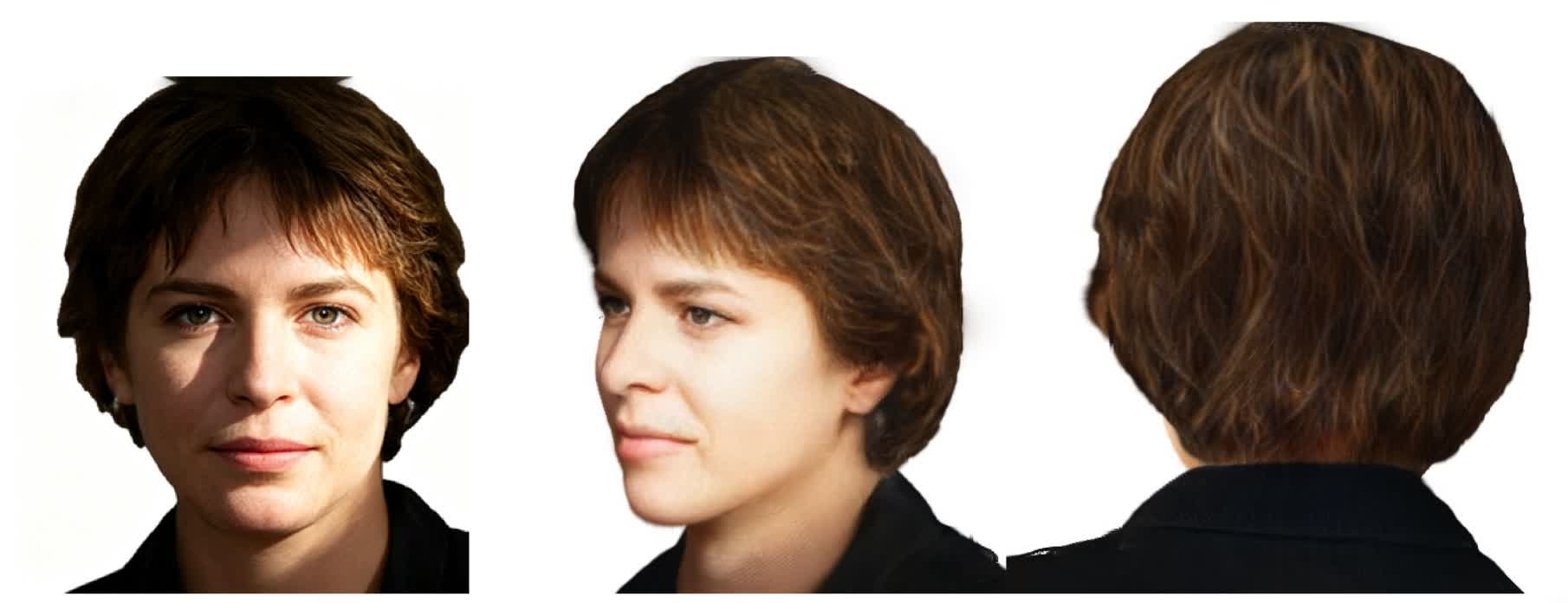}%
    }\hfill
    \small\subfloat[Face tattoo.]{%
        \includegraphics[width=0.49\linewidth]{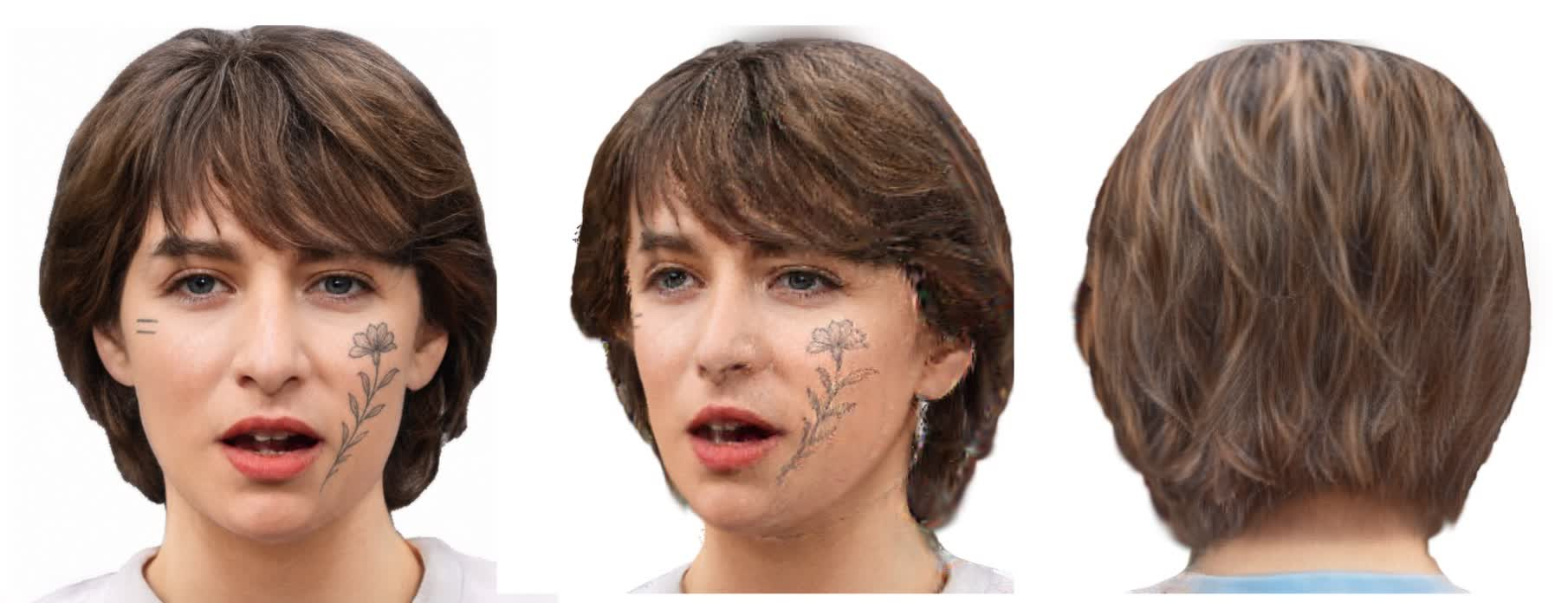}%
    }
    \vspace{-1pt}
    \caption{Out-of-domain qualitative results of PanoLAM under challenging input conditions. Each triplet shows the input and two synthesized novel views.}
    \label{fig:ood}
\end{figure}

\paragraph{\textbf{Qualitative comparisons}}
Fig.~\ref{fig:vis_test} and Fig.~\ref{fig:comp_facelift} illustrate the novel view synthesis results after reconstruction from various methods on the synthesis and out-of-domain real-world images, highlighting the superior performance of PanoLAM. Unlike prior approaches, PanoLAM enhances texture detail, maintains identity fidelity, and ensures multiview consistency. Compared to PanoHead and SphreHead, there are no artifacts in the background as well. PanoLAM reconstructs a head in 0.11 s, compared with 89.4 s for PanoHead-PTI and 96.2 s for SphereHead-PTI (Table~\ref{tab:running_time}), corresponding to approximately $813\times$ and $875\times$ speedups in reconstruction, respectively. These baselines require per-image test-time optimization, whereas PanoLAM uses a single forward pass without requiring an input camera pose. Additional multi-view comparisons are shown in Fig.~\ref{fig:more_cmp_testset}. We also test PanoLAM on out-of-domain challenging cases in Fig.~\ref{fig:ood}, which illustrates its behavior under challenging inputs.

\paragraph{\textbf{Quantitative comparisons}}
\begin{table}[t]
    \caption{Quantitative evaluation on synthesis and real-world testsets.}
    \centering
    \setlength{\tabcolsep}{2.8pt}
    \label{tab:Quant_main_results}
    \scriptsize
    \begin{tabular}{l|ccc|ccc}
        \toprule
        Method & \multicolumn{3}{c|}{Real-world} & \multicolumn{3}{c}{Synthesis} \\
        \cmidrule(lr){2-4} \cmidrule(lr){5-7}
        & PSNR$\uparrow$ & LPIPS$\downarrow$ & SSIM$\uparrow$  & PSNR$\uparrow$ &  LPIPS$\downarrow$ & SSIM$\uparrow$ \\
        \midrule
        PanoHead-PTI     & 16.260 & 0.165 & 0.704  & 24.37 & 0.141 & 0.806 \\
        SphereHead-PTI   & 16.983 & 0.142 & 0.698  & 24.65 & 0.127 & 0.811 \\
        LGM              & 11.472 & 0.444 & 0.572  & 21.33 & 0.350 & 0.765 \\
        LGM-Finetuned & 19.313 & 0.182  & 0.699  & 25.67 & 0.135 & 0.806 \\
        LAM              & 15.627 & 0.245 & 0.647  & 18.72 & 0.157 & 0.713 \\
        LAM-Finetuned & 21.446 & 0.143  & 0.769  & 24.89 & 0.118 & 0.859 \\
        FaceLift & 21.765 & 0.155  & 0.744  & 24.12 & 0.124 & 0.844 \\
        Ours           & \textbf{23.494} & \textbf{0.107} & \textbf{0.793}  & \textbf{27.45} & \textbf{0.093} & \textbf{0.882} \\
        \bottomrule
    \end{tabular}
\end{table}
We benchmark PanoLAM against all baselines on both a held-out synthetic test set and out-of-domain real-world data, evaluating each subject across four viewpoints (front, back, left, and right). As shown in Table~\ref{tab:Quant_main_results}, PanoLAM achieves the best PSNR, SSIM, and LPIPS on the test set, and these gains are obtained in a single forward pass without test-time optimization. To assess generalizability, we further evaluate on 18 real-world identities sampled from VFHQ~\cite{vfhq2022} and Nersemble~\cite{Nersemble_kirschstein2023} following the protocol of GRAM-HD~\cite{GRAM_HD}: for each method, 30 views are synthesized, reconstructed via NeuS2~\cite{neus2}, and scored by PSNR and SSIM. On this out-of-domain benchmark, PanoLAM also consistently outperforms competing methods. In contrast, LAM~\cite{LAM_Sig25}, trained on videos with limited viewing ranges, struggles to reconstruct side and back regions, leading to a pronounced drop in $360^{\circ}$ evaluation. We also fine-tuned baselines, e.g., LGM and LAM, on our proposed datasets in the table; the improved results also validate the effectiveness of the proposed large-scale multi-view dataset.

\begin{table*}[t]
\begin{minipage}[t]{0.30\textwidth}
    \makeatletter\def\@captype{table}\makeatother
    \caption{Running time for 3D avatar generation and rendering.}
    \vspace{-2mm}
    \label{tab:running_time}
    \centering
    \scriptsize
    \setlength{\tabcolsep}{4pt}
    \begin{tabular}{l|cccc}
        \toprule
        Time          & PH-PTI & SH-PTI & LAM            & Ours \\
        \midrule
        Recon.(s)     & 89.4   & 96.2   & 1.4            & \textbf{0.11} \\
        Render(ms)    & 10.29  & 19.68  & \textbf{3.6}   & \textbf{3.6} \\
        \bottomrule
    \end{tabular}
\end{minipage}
\hfill
\begin{minipage}[t]{0.38\textwidth}
    \makeatletter\def\@captype{table}\makeatother
    \caption{Effect of coarse-to-fine generation. OOM: Out of memory (80\, GB).}
    \vspace{-2mm}
    \label{tab:ablation_mem_comp_consumtion}
    \centering
    \scriptsize
    \setlength{\tabcolsep}{4pt}
    \begin{tabular}{l|cccc}
        \toprule
        Metrics                 & D-30K & C2F-30K & D-100K & C2F-100K \\
        \midrule
        Infer.(s)$\downarrow$   & 0.49  & 0.082   & --     & 0.11 \\
        Memory(G)$\downarrow$   & 17.1  & 7.22    & OOM    & 12.7 \\
        \bottomrule
    \end{tabular}
\end{minipage}
\hfill
\begin{minipage}[t]{0.27\textwidth}
    \makeatletter\def\@captype{table}\makeatother
    \caption{Module ablations.}
    \vspace{-2mm}
    \label{tab:module_ablation}
    \centering
    \scriptsize
    \setlength{\tabcolsep}{4pt}
    \begin{tabular}{l|cc}
        \toprule
        Method                & PSNR$\uparrow$ & SSIM$\uparrow$ \\
        \midrule
        Full                  & 23.494 & 0.793 \\
        w/o C2F               & 22.312 & 0.731 \\
        w/o ST ray agg.       & 22.156 & 0.715 \\
        w/o ST & 21.814  & 0.712 \\
        2K subj.              & 19.391 & 0.699 \\
        \bottomrule
    \end{tabular}
\end{minipage}
\end{table*}

\paragraph{\textbf{Comparisons of Reconstruction and Rendering Efficiency}}
Table.~\ref{tab:running_time} shows the comparative running time evaluation of all methods on an A100 GPU. Benefiting from our coarse-to-fine and feedforward framework design, our approach achieves a fast reconstruction speed. The Gaussian Splatting we chose as the 3D representation also enables fast rendering speeds compared to existing techniques.

\subsection{Ablation Studies}
To distinguish architectural effects from dataset scale, the component ablations in Table~\ref{tab:module_ablation} use the same training dataset as the full model, whereas the \textit{2K subj.} variant retains the full architecture and reduces the number of training identities. The former evaluates the proposed components under fixed training data; the latter evaluates the effect of training-set scale.

\paragraph{\textbf{Effect of Coarse-to-fine Gaussian Point Generation}}
\begin{figure}[t]
    \centering
    \includegraphics[width=\linewidth]{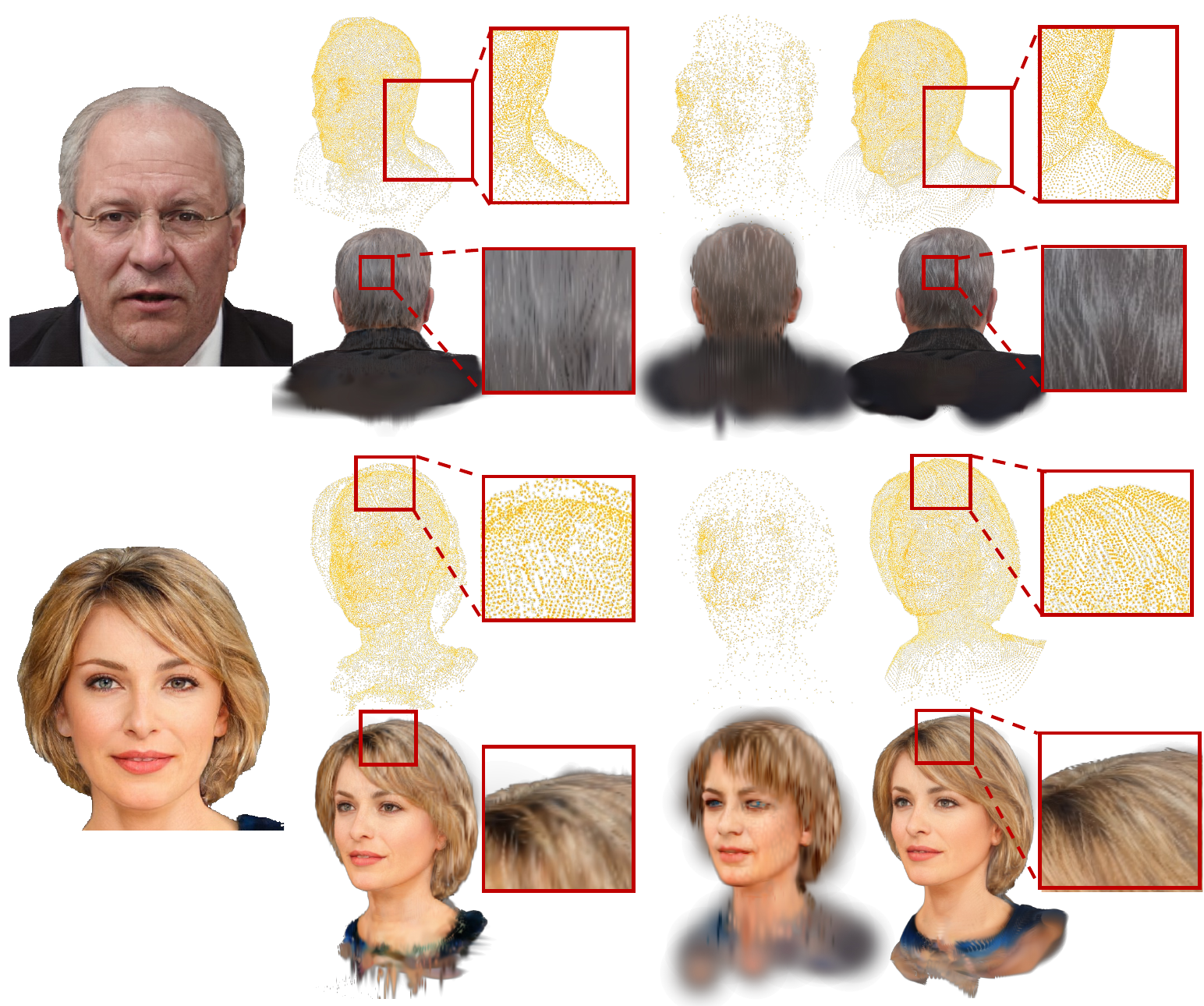}
    \scriptsize \leftline{~~~\quad~~~Input~\qquad\quad~W/o coarse-to-fine~~~~~~~Coarse~~~~~~~~~~Fine~(Ours)}
    \vspace{-10pt}
    \caption{Effect of coarse-to-fine generation. Without the coarse-to-fine strategy, the dense Gaussians are distributed on multiple layers, leading to degradation of rendering quality. Zoom in for details.
    }
    \label{fig:ablation_coarse2fine}
    \vspace{-4mm}
\end{figure}
Fig.~\ref{fig:ablation_coarse2fine} shows the effect of our coarse-to-fine Gaussian generation strategy. The sparse points are easier to optimize and can be easily deformed to carve the shape of the person, but they cannot model texture details like hair strands. 
By densifying the points with a topology prior in the FLAME mesh, our coarse-to-fine model can learn more detailed textures. 
In contrast, directly regressing the dense point cloud results in insufficient optimization of Gaussian points. Many points are not well optimized for deformation and remain on the FLAME surface, which are invisible and wasted, causing dual-layer geometry in the hair region as well. The quantitative results in Tab.~\ref{tab:module_ablation} also support the effectiveness; without our coarse-to-fine training scheme (denoted as w/o C2F in the table), the performance drops. Moreover, as shown in Table~\ref{tab:ablation_mem_comp_consumtion}, our coarse-to-fine strategy saves computation and GPU memory consumption since fewer points are needed for stacked cross-attention modules. The network forward time and training GPU consumption are also reduced with our coarse-to-fine training strategy.

\begin{figure}[t]
    \centering
    \includegraphics[width=\linewidth]{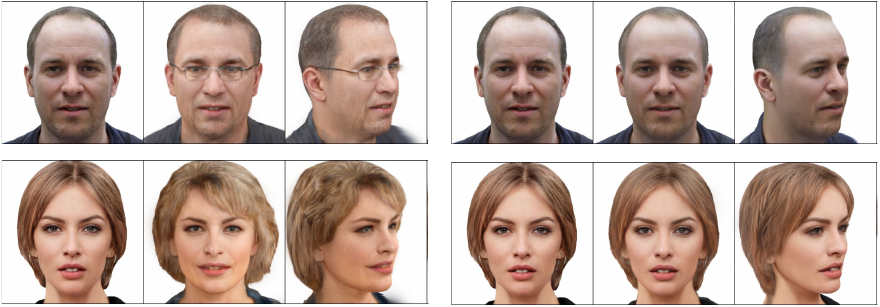}
    \scriptsize \leftline{~~~~~~Input~~~~~\qquad~~~~~~2K Subj.~\quad\qquad\qquad~~~~Input~~~\qquad~~~~165K Subj.}
    \vspace{-4mm}
    \caption{Comparison of our network trained on different numbers of subjects. The model trained on a small number of subjects cannot generalize well to unseen identities.
    }
    \label{fig:ablation_sub_number}
    \vspace{-2mm}
\end{figure}
\paragraph{\textbf{Effect of Different Dataset Scale}}
Previous 3D head datasets lack subject diversity, as shown in Table.~\ref{tab:data}. We ablate the effect of different numbers of subjects in Fig.~\ref{fig:ablation_sub_number} and Table~\ref{tab:module_ablation} (2K subj.). The model trained on a limited number of subjects cannot generalize well to unseen subjects, causing performance drops visually and quantitatively. This shows the effectiveness of our proposed large-scale, diverse dataset for network generalizability.

\paragraph{\textbf{Effect of Different Spherical Triplane Query Strategy.}}

\begin{figure}[t]
    \centering
    \includegraphics[width=\linewidth]{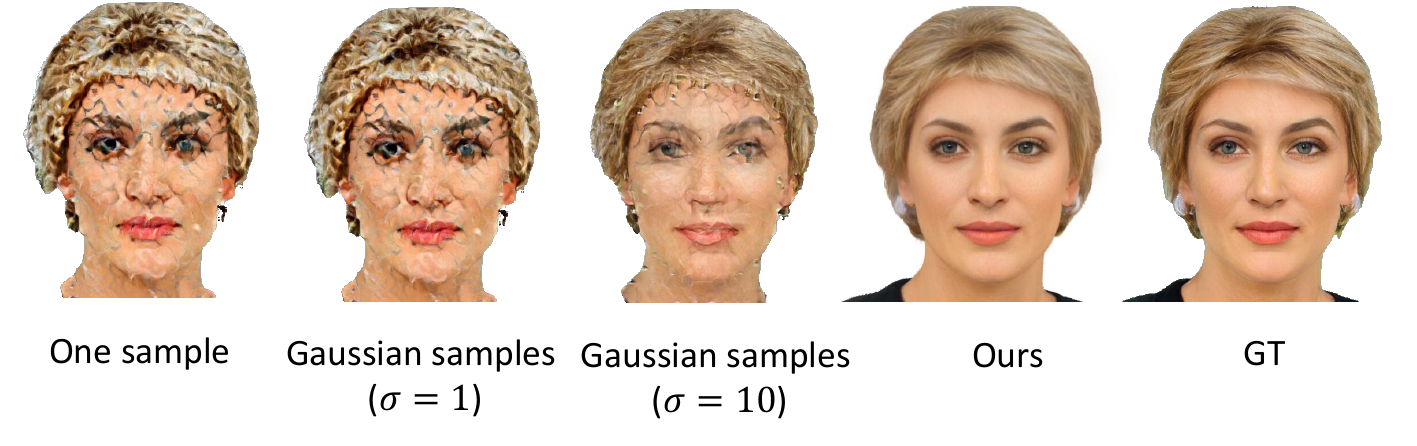}
    \vspace{-10pt}
    \caption{Comparisons on different spherical triplane feature query and aggregation strategies. For each point, sampling only one sample of features from the spherical triplane cannot fully fetch all valuable features, resulting in a collapsed rendered image. Aggregating 32 points with Gaussian distribution weights gets better results, but still blurry. Our MLPs learn to aggregate features as the original NeRF and get quite better results.
    }
    \label{fig:ablation_vis_triplane_feat_agg}
    \vspace{-2mm}
\end{figure}

\begin{figure}[t]
    \centering
    \includegraphics[width=\linewidth]{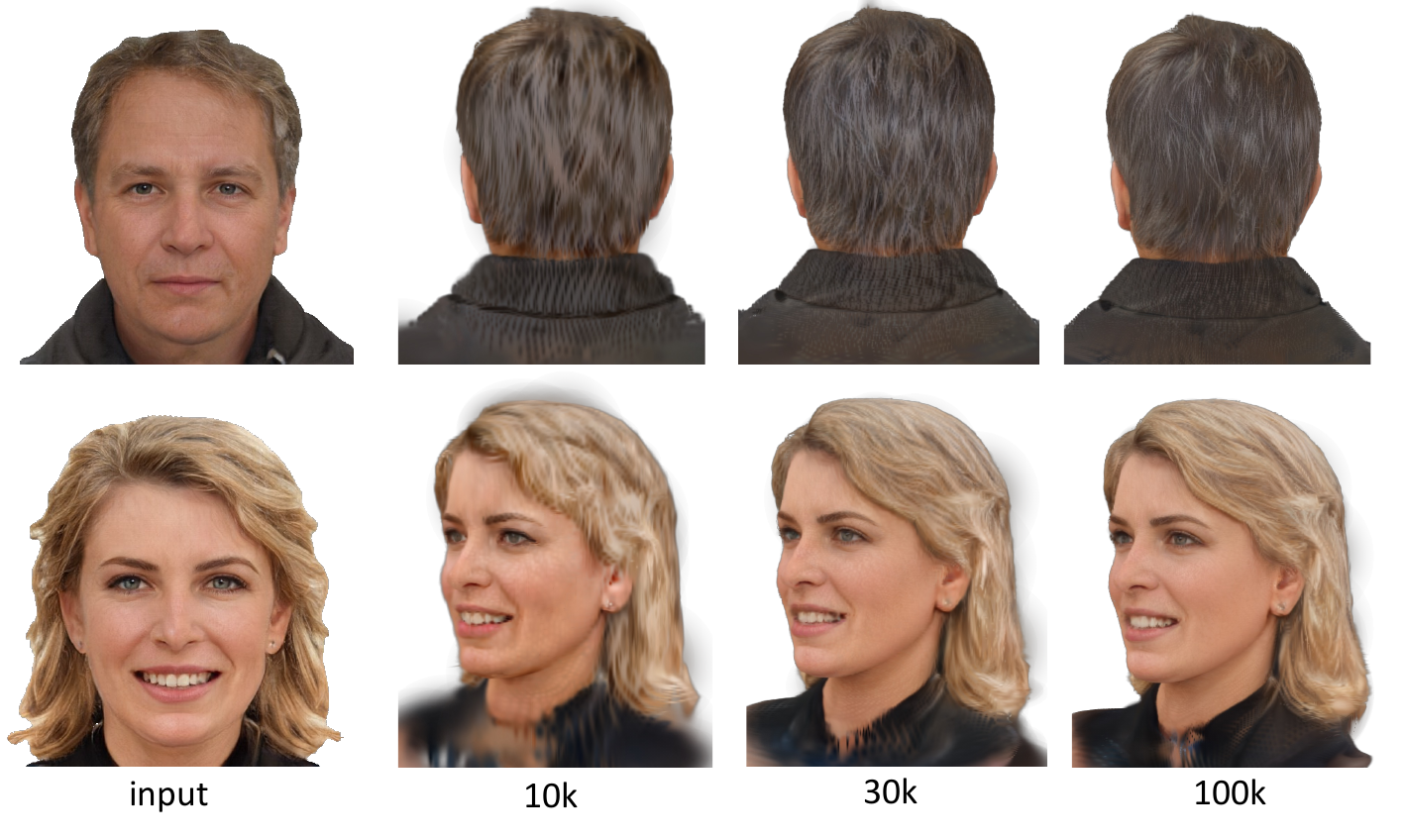}
    \caption{Comparison of different numbers of Gaussian points. A small number of Gaussian points cannot model texture details like hair strands and teeth. More high-fidelity Gaussian avatars are reconstructed with an increasing number of Gaussian points. Zoom in to see more details.}
    \label{fig:ablation_GS_number}
\end{figure}

\begin{table}[t]
    \caption{Quantitative comparison of different numbers of Gaussian points on the test set.}
    \centering
    \setlength{\tabcolsep}{2.8pt}
    \label{tab:Quant_ablation_points}
    \small

    \begin{tabular}{l|ccccc}
        \toprule
        Method & PSNR$\uparrow$ & LPIPS$\downarrow$ & SSIM$\uparrow$ & Infer.(s)$\downarrow$ & Memory(G)$\downarrow$ \\
        \midrule
        10K  & 22.22 & 0.203 & 0.738 & 0.074 & 5.65 \\
        30K  & 26.42 & 0.134 & 0.835 & 0.082 & 7.22 \\
        100K & 27.45 & 0.093 & 0.882 & 0.110 & 12.70 \\
        \bottomrule
    \end{tabular}
\end{table}

In Fig.~\ref{fig:ablation_vis_triplane_feat_agg}, we show the results of different query strategies from the spherical triplane. All strategies use the same feature renderer and frozen decoder from SphereHead for target image decoding. As shown in the figure, directly using the densified point cloud to query a single layer of features from the spherical triplane, as in~\cite{TripMeetGS_ZouYGLLCZ24}, cannot fetch all valuable features, and the rendered images contain artifacts due to missing information. We then add 4 virtual cameras and cast rays from the camera center to each point and sample 32 layers of features near the point surface for feature aggregation. 

We set the margin between each sampled point to be the same as the fine ray marching stage in SphereHead. We try to assign manual weights sampled from a normal distribution centered at the selected point with standard deviations of 1 and 10, and aggregate sampled features with a weighted sum. As shown in the middle part of Fig.~\ref{fig:ablation_vis_triplane_feat_agg}, the results look better but are still blurry. Our final solution utilizes learnable MLPs to learn the aggregation function for the NeRF-based rendering. As shown in the figure, ours gets the best results, showing the effectiveness of the proposed aggregation strategy. In Table.~\ref{tab:module_ablation}, we also ablate the model without our proposed ray-based feature aggregation from spherical triplane (denoted w/o ST ray agg. in the table), and, even further, the model without spherical triplane branch (denoted w/o ST in the table). The quantitative results drop and prove the effectiveness of the proposed Spherical triplane branch and the ray-based feature aggregation mechanism.

\paragraph{\textbf{Effect of Different Number of Gaussian Points.}}
We ablate the effect of different numbers of Gaussian points, controlled by different face subdivision threshold $\lambda_{area}$, with our coarse-to-fine strategy in Fig.~\ref{fig:ablation_GS_number} and Table~\ref{tab:Quant_ablation_points}. As shown in the figure, a small number of Gaussian points cannot model texture details like hair strands and teeth, leading to blurry results in these regions. In contrast, more high-fidelity Gaussian avatars are reconstructed with an increasing number of Gaussian points. However, directly increasing the number of Gaussian points causes the problem of insufficient optimization of Gaussians and out-of-memory issues, as shown in Table~\ref{tab:ablation_mem_comp_consumtion}. In contrast, as shown in Table~\ref{tab:Quant_ablation_points}, our proposed coarse-to-fine strategy not only enables higher fidelity results, but also low memory consumption. This further validates the effectiveness of our proposed coarse-to-fine training strategy.

%% file: sections/05_conclusion.tex
\section{Conclusions and Limitations}

In this work, we present a novel feedforward large avatar model for Gaussian full-head synthesis from a single unposed image. To resolve the problem of lacking 3D assets for network training, we develop a large-scale, diverse 3D avatar dataset from trained 3D GANs. We also propose a coarse-to-fine mechanism with a network-free point cloud densification strategy for efficient reconstruction of a high-fidelity Gaussian full-head from a single image. We also introduce a dual-branch approach to distill knowledge from the spherical triplane prior and improve the reconstruction effectiveness. We analyze the feature sampling mechanism in spherical triplane and propose an efficient aggregation mechanism. Experimental results validate the effectiveness of our approach.

\textbf{Limitations.} 
Our training data are sampled from pretrained 3D GAN priors, so the proposed model inevitably inherits their distribution biases. As a result, performance may degrade on underrepresented domains, such as certain demographics and stylized/cartoon identities. Also, in out-of-domain heavy occlusion cases, e.g., face occluded by hand in Fig.~\ref{fig:ood}, artifacts may appear near the occluded facial region because the visible identity cues are corrupted by an object that is not part of the head. In addition, our current framework focuses on static full-head reconstruction and does not model animation; extending the method with more diverse priors and animatable dynamics is an important direction for future work.

\textbf{Ethics Statement.} The generation of realistic avatars from images raises important ethical considerations, especially concerning privacy, consent, and the risks associated with the misuse of technology, such as the creation of deep fakes. It is essential to prioritize responsible use, ensuring adherence to ethical guidelines and standards to mitigate potential negative impacts.